\documentclass[letterpaper,times]{sae}

\usepackage[svgnames]{xcolor}
\definecolor{SAEblue}{RGB}{0,90,150}
\raggedright

\usepackage[justification=justified, singlelinecheck=false]{caption}  
\DeclareCaptionFont{eightpt}{\fontsize{8pt}{11pt}\selectfont #1}
\usepackage{lineno}
\usepackage[utf8]{inputenc}
\usepackage{amssymb}
\usepackage{hyperref}

\usepackage{array}
\newcolumntype{L}[1]{>{\raggedright\let\newline\\\arraybackslash\hspace{0pt}}p{#1}}
\newcolumntype{C}[1]{>{\centering\let\newline\\\arraybackslash\hspace{0pt}}p{#1}}
\newcolumntype{R}[1]{>{\raggedleft\let\newline\\\arraybackslash\hspace{0pt}}p{#1}}

\usepackage{graphicx}
\usepackage{multirow}
\usepackage{amsmath}

\newcommand{\ignore}[1]{}

\usepackage{nameref}
\usepackage{booktabs}

\makeatletter
\def\@seccntformat#1{%
  \expandafter\csname c@#1\endcsname\c@section
  }
\makeatother

\usepackage{subfigure}
\usepackage{multimedia}

\PaperTitle{AI-Driven Repair of Missing Automotive Components}
\usepackage{calc} 

\makeatletter 
\renewcommand\@biblabel[1]{#1. } 
\makeatother

\AddAuthor{Kazim Abbas}{University of Michigan-Dearborn}
\AddAuthor{Zheng Liu}{University of Michigan-Dearborn}
\PaperNumber{27IDM-0028}
\SAECopyright{2026}

\usepackage{placeins}
\usepackage{cuted}
\begin{document}
\maketitle
\section*{Abstract}

Due to supply constraints, replacing damaged automotive components can be expensive and time-consuming, while replacing the entire component can be wasteful when much of it remains usable. This paper presents a pipeline to investigate whether missing geometry can be reconstructed as a repair patch using paired observations of the intact models and synthetically damaged components. Noise-free color, depth, and silhouette images were produced using the camera viewpoints. Depth differences were then used to identify the potential missing regions, while multi-view voxel reconstruction was used to estimate the missing portion. A pretrained DINOv2 model was also added to validate missing candidates and determine whether pretrained visual features could improve reconstruction. We tested the pipeline on 20 automotive CAD models, each with a synthetically removed portion. Each model was tested under four different conditions using two reconstruction methods for a total of 160 trials. The predictions were compared with the missing geometry of the unchanged ground-truth using predefined thresholds of the voxel intersection-over-union $\geq 0.85$ and the absolute relative volume error $\leq 15\%$. Of the 160 experimental trials, 109 of them satisfied both thresholds. Geometry-only reconstruction passed 16 of 20 components with eight views but only 10 of 20 with four adjacent views. Further analysis showed that 24 out of the 37 failed reconstructions could be categorized as unsuccessful because extra material was incorrectly added to areas that should have been empty. The implementation of DINOv2 into the pipeline did not improve the overall reconstruction accuracy and made prediction generation slower. For 13 verified tests, printing only the repaired sections instead of the entire parts reduced filament use by 93.08\% and printing time by 89.39\% based on the median results. These findings demonstrate the possibilities and geometric limitations of repair-only reconstruction under controlled synthetic conditions, as well as potential use cases for artificial intelligence.

\section{Introduction}

Replacement of damaged automotive components can be expensive and time-consuming, particularly when replacement parts are difficult to obtain or have been discontinued. Research into additive manufacturing has explored on-demand production as an alternative approach for automotive spare-parts supply \cite{source1}. Previous research has also showcased workflows that use existing technical drawings to create models for reproducing unavailable automotive spare parts through additive manufacturing \cite{source6}.

While full part replacement is possible, there are cases where a large portion of the original component remains usable and the damage is limited to a section of the model. For a scenario like that, producing a repair patch for the specific damaged section would reduce material usage and printing time compared to reproducing the entire component. Creating repair patches often follows similar methodologies. To construct a repair patch, you would take information from the original CAD model and the damaged state to compare them. The goal would be to isolate the missing geometry as accurately as possible without including any of the intact portions. Previous research has investigated several ways to accurately complete the steps. Among these methods, researchers have tried generating repair pieces from aligned 3D scans, comparing intact and damaged voxel models, and reconstructing damaged turbine components to identify missing material \cite{source2, source3, source7, source8}. Learning-based approaches have also been explored by estimating missing geometry from single images, incomplete 3D scans, and fractured objects \cite{source4, source9, source10}. Additionally, pretrained computer vision features have been investigated for industrial anomaly detection, suggesting a possible approach for identifying visual differences between intact and damaged components \cite{source5}. 

Despite these developments, there are fundamental challenges that still arise from partial reconstruction, especially relating to limited visual information. Research into silhouette-based reconstruction has demonstrated that certain geometric features cannot always be recovered from visible outlines alone \cite{source13}. Components with hollow openings or internal cavities are difficult to reconstruct because the available cameras cannot always observe those regions. A camera may capture the exterior portions of a component while missing important details inside it. Using viewpoints distributed around the component could reveal information that several closely positioned cameras would miss but even then, there is no guarantee that they capture every missing detail. This raises a question: how can missing geometry be accurately reconstructed when components contain complex internal structures or regions that are difficult to observe?

This paper investigates a CAD-guided approach for reconstructing missing geometry in automotive components. A controlled virtual study allows for the full volume of the model to be known before the start of any reconstruction. This setup allowed camera coverage and RGB brightness to be varied while maintaining noise-free depth maps and known camera positions. We can therefore study the effects of various conditions without introducing error produced from real world scanning  

\begin{figure*}[t]
    \centering
    \includegraphics[width=0.98\textwidth]{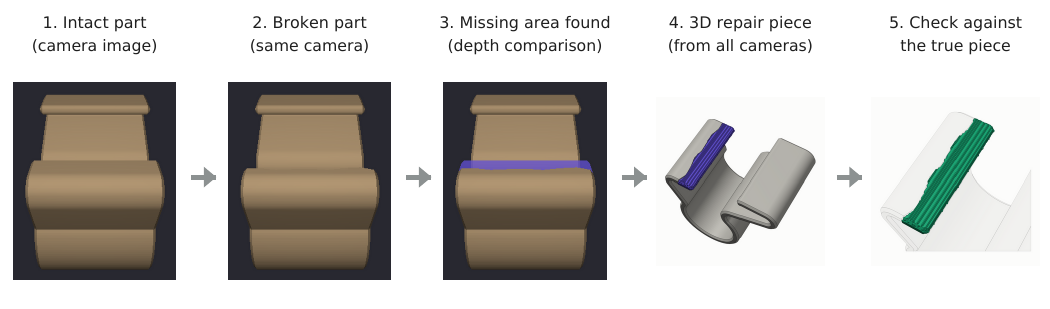}
    \caption{Overview of the CAD-guided reconstruction pipeline for generating the repair patch}
    \label{fig:pipeline}
\end{figure*}

The idea of the process is to take an automotive CAD model and generate a synthetically damaged counterpart by removing a section. Both versions are then rendered from matching viewpoints to produce corresponding RGB image pairs and depth maps. Differences between the depth maps identify potential missing regions, while a pretrained DINOv2 model is used to confirm or reject these regions. Finally, depth information from multiple viewpoints is combined using voxel-based reconstruction to recover the missing volume and generate a repair-only stereo-lithography mesh. The pipeline was tested on 20 automotive CAD models under four different viewing conditions and two reconstruction methods, resulting in 160 trials. These tests measured reconstruction accuracy, the effects of limited viewing angles, and whether DINOv2 improved the results.

Reconstruction accuracy was evaluated by comparing each predicted repair against the protected ground-truth missing geometry using voxel intersection-over-union (IoU) and relative volume error. An additional evaluation used PrusaSlicer to compare slicer-estimated filament consumption and printing time between complete components and their corresponding repair sections. The result was 13 verified paired comparisons under controlled printing settings. These experiments judge the potential production efficiency of the reconstruction process under controlled conditions.

\section{Methodology}

Fig.~\ref{fig:pipeline} shows a simplified version of the overarching methodology. The process generates a repair patch from paired observations. The intact and damaged components are rendered from matching camera viewpoints, and their depth information is compared to identify possible missing regions. Information from multiple viewpoints is then combined to reconstruct the three-dimensional repair patch. Finally, the generated repair is compared with the actual missing geometry to measure its accuracy.

\subsection{2.1. CAD Preparation and Damage Localization}
\label{sec:cad-preparation}
Each of the selected automotive components is represented via an intact CAD solid $N$ in a three dimensional coordinate space $W$, measured in millimeters. The implementation uses a box-shaped cutter $K$ aligned with the coordinate axes, with an uneven, jagged surface representing the crack to produce a damaged solid $D = N\setminus K$ and the true missing geometry $T = N \cap K$. The intact and damaged components are then rendered from eight camera angles oriented at $0^\circ$, $90^\circ$, $180^\circ$, and $270^\circ$ at elevations of $\pm 30^\circ$. Each viewpoint produces a $672 \times 504$ RGB image, a metric depth map, and a silhouette frame. During reconstruction, the predictor only has access to these observations from the camera information, while the original CAD solids and ground truth missing geometry remain locked.

For each camera viewpoint, the reference and damaged depth maps are compared to generate a candidate mask. The following indicators are used to mark a pixel as missing:
(a): the reference ray intersects the object at some pixel (p) while the damaged ray does not.
(b): both rays intersect the object at pixel $p$, but the damaged surface is farther away than the reference surface by more than the predefined threshold $\tau = 0.02\,\mathrm{mm}$.
These conditions are used to generate a candidate mask $C_v$ for each viewpoint, identifying pixels where material may have been synthetically removed.

\subsection{2.2. Multi-View Voxel Reconstruction}
\label{sec:voxel-reconstruction}

After identifying the potential candidates for missing regions, the system uses a three-dimensional voxel grid to reconstruct the missing portion \cite{source11}. A voxel is a small cube representing a portion of 3D space. The size of the voxel is determined by using the automotive component's feature thickness and camera resolution, with sizes ranging from 0.13 to 0.46 mm in our experiments. The system then uses depth carving to estimate the shapes of the intact and damaged components separately. Depth carving functions by eliminating voxels in areas where camera information tells the system that the space is empty \cite{source12}. The voxel-reconstructed original and damaged components are compared in an aligned 3D grid to identify which voxels appear in the intact but not in the damaged reconstruction. Each missing voxel candidate must also be supported by at least two camera viewpoints. The process can be expressed by Eq. \eqref{eq:reconstruction}.

\begin{equation}
V_R = (V_N \setminus V_D) \cap S_2
\label{eq:reconstruction}
\end{equation}
where $V_R$ represents the predicted repair voxel, $V_N$ represents the estimated intact volume, $V_D$ represents the estimated damaged volume, and $S_2$ represents voxels supported as missing by at least two camera viewpoints.

\subsection{2.3. Repair Generation and DINOv2-Assisted Variant}
\label{sec:repair-generation}

Upon the reconstruction of the missing region, the system confirms that the predicted voxels generate a suitable repair patch. It starts by grouping voxels that share a face using six-neighbor connectivity. Small disconnected groups are removed if they contain fewer than 27 voxels or 1\% of the total predicted voxel repairs, whichever is larger. The repair portion is rejected if more than one of the connected groups remains or if more than 2\% of the original predicted voxels were removed. The remaining voxels are then converted into a triangular surface mesh using the marching cubes algorithm. Before the last step of exporting the repair as an STL file is carried out, the mesh must be watertight, have consistently oriented faces, and contain a positive enclosed volume. Predictions that fail these criteria are rejected.

\begin{figure*}[!t]
    \centering
    \includegraphics[width=0.9\textwidth]{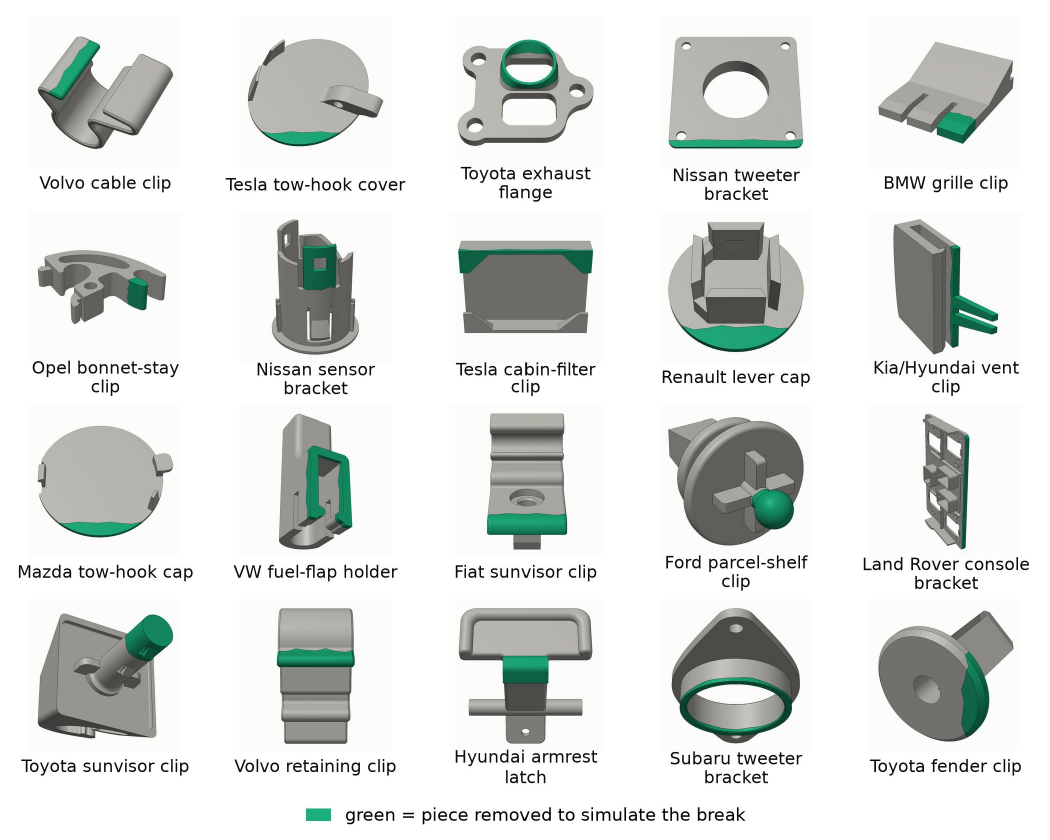}
    \caption{The 20 automotive CAD components used in the evaluation. Green identifies the synthetically removed region for each component.}
    \label{fig:component_examples}
\end{figure*}

In addition to the first geometry-only method, a second variant was implemented using a pretrained DINOv2 model \cite{source14}. The goal was to examine if visual features could improve reconstruction. The model compared features from both the intact and damaged RGB images to identify differences in appearance. A visual-change score would then be calculated using Eq. \eqref{eq:dinov2-score}.

\begin{equation}
S = 1-\cos(f_N,f_D)
\label{eq:dinov2-score}
\end{equation}
where $f_N$ and $f_D$ represent features from the intact and damaged images. The system sets a change threshold using the 98th percentile of scores from an image region outside the depth-based missing candidates. If the region contained fewer than 50 image-patch areas, the viewpoint provided no acceptable candidates. Otherwise, a missing-region candidate was accepted when at least 25\% of its pixels met the visual-change threshold. These accepted regions would then be used in the same voxel reconstruction process described in Section~\ref{sec:voxel-reconstruction}.

\section{Experimental Evaluation}

The evaluation for this experiment used 20 unique automotive CAD components obtained from publicly available websites. These included 10 clips, 4 brackets, 3 cover caps, 2 holders/latches, and 1 flange. Each component consisted of one single solid that successfully passed the geometric preparation tests. After passing the preparation tests, portions of the model were synthetically removed from each component using a box cutter with a jagged surface. The sections removed represent anywhere between $3.4\%-5.4\%$ of the original volumes of the components. The same synthetic damage was maintained for each component throughout all experimental conditions.

Fig.~\ref{fig:component_examples} shows the 20 automotive components and their synthetically removed regions, highlighted in green. The components differ in their geometric features, including openings, thin sections, and protrusions, which can affect how much of the damaged region is visible from the available viewpoints.

Four observation conditions were used to investigate the effects of visual coverage and changes in image appearance. The first condition used all eight camera viewpoints. The second used four viewpoints that were evenly spread around the component, while the third used four adjacent camera viewpoints covering a $90^\circ$ range. The fourth condition used all eight cameras but reduced the brightness of the damaged RGB images to $65\%$ of the original brightness. The depth maps remained unchanged for entire experimental process. Each condition was tested with both a geometry-only method and a DINOv2 assisted method, which resulted in 160 experimental trials across the 20 components.

The predicted repairs were compared against the original missing geometry. Accuracy was measured using voxel intersection-over-union, which measured the overlap between the predicted repair and the actual missing piece. Relative volume error was also calculated to measure whether the predicted volume was too large or too small. 

\begin{equation}
\mathrm{IoU} = \frac{|R \cap T|}{|R \cup T|},
\qquad
e_V = \frac{V(R)-V(T)}{V(T)}
\label{eq:evaluation}
\end{equation}

\begin{figure*}[t]
    \centering
    \includegraphics[width=0.88\textwidth]{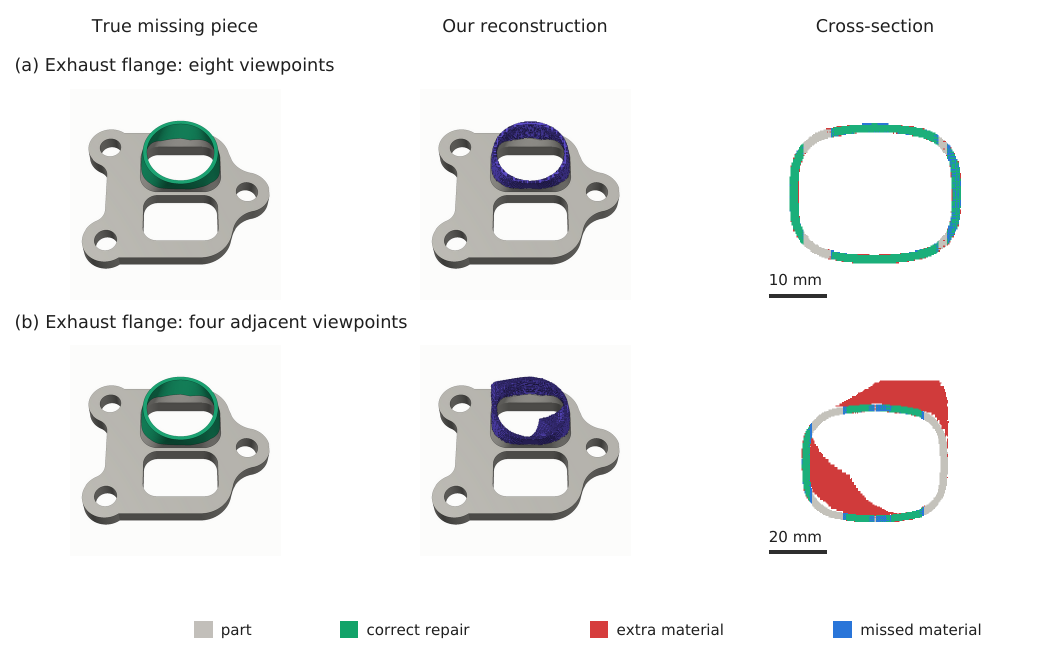}
    \caption{Comparison of exhaust flange reconstruction under different camera conditions. (a) Eight viewpoints (IoU = 0.898). (b) Four adjacent viewpoints (IoU = 0.369). Red indicates extra material; blue indicates missed material.}
    \label{fig:failure}
\end{figure*}

In Eq. \eqref{eq:evaluation}, $R$ represents the predicted repair, $T$ represents the untouched original missing geometry, and $V$ represents the volume. A reconstruction passed only if the intersection-over-union was at least 0.85 and its volume error was no greater than $15\%$. These thresholds were defined prior to the experiment. Rejected predictions were considered failures and were not assigned an intersection-over-union score.

For our final evaluation, PrusaSlicer 2.9.6 was also used to estimate the material consumption and printing time of repair-only sections compared with complete components. The method PrusaSlicer uses for estimating the material consumption is by generating a filament path from the sliced geometry. The required filament is determined by the volume assigned to those paths. Printing time is estimated via the customization movement speed and acceleration based on what the user choose. Both of these are calculated separately from the original components filament usage and printing time. In our experiment, 16 geometry-only repairs were chosen under the criteria that they passed the eight-view accuracy requirements. Both versions of each component were processed using the same Prusa MK4 printer profile, PLA material, layer height, two perimeters, $15\%$ grid infill, automatic supports, and brim settings. Original orientations were also preserved. The percentage reduction was calculated as:

\begin{equation}
\Delta_q =
100\left(1-\frac{q_{\mathrm{repair}}}
{q_{\mathrm{full}}}\right)\%
\label{eq:slicing-reduction}
\end{equation}
where $q_{\mathrm{repair}}$ and $q_{\mathrm{full}}$ represent the estimated material usage or printing time for the repair and complete component.
Out of the 16 chosen models, 13 of them passed the slicing verification tests and were included within the final analysis. The remaining three components were excluded because of slicing verification issues. The Mazda cover failed to generate first layer filament paths which prevented successful G-code exports. For the Volkswagen flap holder and Ford shelf clip, PrusaSlicer automatically removed degenerate triangles from the complete-component meshes. This violated the requirement to preserve the original geometry during comparison. None of the components were physically printed during testing.

\section{Results and Limitations}

\begin{table}[h]
\centering
\caption{Reconstruction outcomes across all 160 trials.}
\label{tab:reconstruction}
\small
\renewcommand{\arraystretch}{1.15}
\setlength{\tabcolsep}{3pt}
\begin{tabular}{@{}llrrr@{}}
\toprule
Views & Method & Pass & Fail & Refused \\
\midrule
8          & Geometry & 16 & 3 & 1 \\
8          & DINOv2   & 16 & 3 & 1 \\
4 spread   & Geometry & 13 & 6 & 1 \\
4 spread   & DINOv2   & 12 & 4 & 4 \\
4 adjacent & Geometry & 10 & 8 & 2 \\
4 adjacent & DINOv2   & 10 & 7 & 3 \\
8 darkened & Geometry & 16 & 3 & 1 \\
8 darkened & DINOv2   & 16 & 3 & 1 \\
\midrule
\multicolumn{2}{l}{Total} & 109 & 37 & 14 \\
\bottomrule
\end{tabular}
\end{table}

Table~\ref{tab:reconstruction} summarizes the reconstruction outcomes across all 160 trials. Failed trials produced repair patches that did not meet one or both accuracy requirements, while refused trials did not produce an accepted repair because the pipeline's reconstruction or validity checks were not satisfied. The median voxel intersection-over-union among passing trials was approximately 0.902.

The geometry-only method passed 16/20 components when eight camera viewpoints were available. This decreased to 13/20 with four camera viewpoints and 10/20 using the four adjacent viewpoints. The results suggest that reduced camera coverage negatively affects the reconstruction process.

\begin{table*}[t]
\centering
\caption{Per-component voxel IoU (\%) for geometry-only reconstruction across camera conditions.}
\label{tab:per_component_iou}
\small
\setlength{\tabcolsep}{9pt}
\renewcommand{\arraystretch}{1.06}
\begin{tabular}{@{}lcccc@{}}
\toprule
Component & 8 views & 4 spread & 4 adjacent & 8 darkened \\
\midrule
Ford parcel-shelf clip & \textbf{94} & \textbf{92} & \textbf{93} & \textbf{94} \\
Hyundai armrest latch & \textbf{93} & \textbf{86} & \textbf{92} & \textbf{93} \\
Kia/Hyundai vent clip & \textbf{91} & \textbf{89} & \textbf{90} & \textbf{91} \\
Mazda tow-hook cap & \textbf{93} & \textbf{89} & \textbf{92} & \textbf{93} \\
Renault lever cap & \textbf{97} & \textbf{93} & \textbf{97} & \textbf{97} \\
Tesla cabin-filter clip & \textbf{91} & \textbf{87} & \textbf{88} & \textbf{91} \\
Tesla tow-hook cover & \textbf{92} & \textbf{90} & \textbf{92} & \textbf{92} \\
Volvo retaining clip & \textbf{89} & \textbf{87} & \textbf{88} & \textbf{89} \\
BMW grille clip & \textbf{93} & \textbf{89} & 73 & \textbf{93} \\
Nissan sensor bracket & \textbf{92} & \textbf{91} & \textemdash & \textbf{92} \\
Nissan tweeter bracket & \textbf{86} & 84 & \textbf{87} & \textbf{86} \\
Toyota exhaust flange & \textbf{90} & \textbf{88} & 37 & \textbf{90} \\
Toyota fender clip & \textbf{89} & \textbf{86} & 80 & \textbf{89} \\
Fiat sunvisor clip & \textbf{86} & 79 & 67 & \textbf{86} \\
Subaru tweeter bracket & \textbf{89} & 84 & \textemdash & \textbf{89} \\
VW fuel-flap holder & \textbf{88} & 80 & 83 & \textbf{88} \\
Volvo cable clip & \textemdash & \textbf{93} & \textbf{93} & \textemdash \\
Land Rover console bracket & 79 & 82 & 84 & 79 \\
Opel bonnet-stay clip & 80 & 75 & 40 & 80 \\
Toyota sunvisor clip & 69 & \textemdash & 69 & 69 \\
\bottomrule
\end{tabular}
\end{table*}

\begin{figure*}[t]
    \centering
    \includegraphics[width=0.82\textwidth]{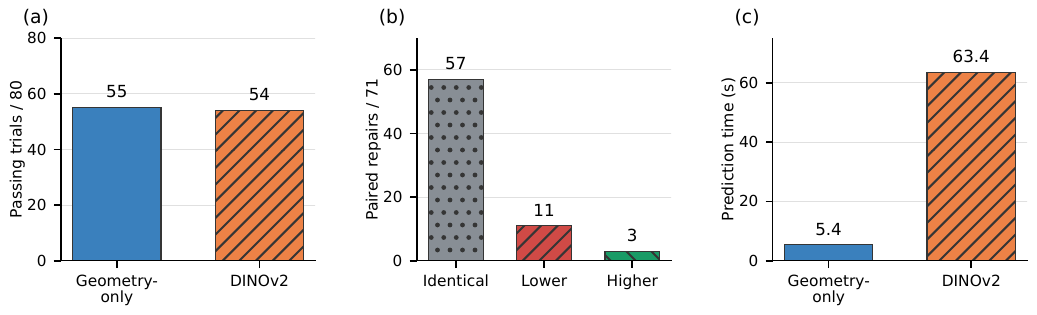}
    \caption{Comparison of geometry-only and DINOv2-assisted reconstruction. (a) Passing trials. (b) Accuracy differences among paired repairs. (c) Median prediction-stage runtime.}
    \label{fig:dinov2_comparison}
\end{figure*}

Table~\ref{tab:per_component_iou} provides the reconstruction accuracy for each component under the four observation conditions. The results show that camera coverage did not affect every component equally. For example, the Toyota exhaust flange's IoU decreased from approximately 0.90 with eight viewpoints to 0.37 with four adjacent viewpoints, while the Ford parcel-shelf clip maintained an IoU above 0.90 under all four conditions.

Examining the failures showed that many unsuccessful reconstructions failed because they added material to regions that should have been empty. About 24 of the 37 inaccurate repairs were dominated by this problem. Fig.~\ref{fig:failure}(a) shows the exhaust flange using eight viewpoints, which achieved an intersection-over-union of 0.898. Fig.~\ref{fig:failure}(b) shows the same flange using four adjacent viewpoints, which achieved an intersection-over-union of 0.369. The three columns show the actual missing section, the predicted repair, and a cross-sectional comparison. The red regions show where the reconstruction added material inside the hollow opening of the flange. This occurred because the limited viewpoints did not provide enough information to distinguish some hidden empty space from missing material.

The geometry method of reconstruction passed 55/80 trials, compared to the DINOv2 assisted method which only passed 54/80. Of the 80 paired trials, both reconstrution methods successfully generated a repair Stereo lithography file for 71 of the cases. From the 71 cases, 57 of them produced identical Stereo lithography files, which indicated that DINOv2 did not change the final reconstrcuted geometry in most cases. The remaining 14 cases were split into two categories where 11 of the cases DINOv2 produced were less accurate and 3 of them were more accurate, as shown in Fig.~\ref{fig:dinov2_comparison}(b). The DINOv2 method did not convert any of the failing cases into passing results, while one of the previously passing cases ended up failing. The median prediction stage runtime increased from 5.4 seconds to 63.4 seconds with DINOv2. Both methods were run on a CPU without any GPU acceleration. The time increase was also for the prediction stage and not the entire pipeline. The results obtained from the experiments showed that a pretrained visual feature filter did not improve the passing performance in most cases. A possible explanation for these results is that the DINOv2 filtering stage could reject missing-region candidates that were correctly identified through depth comparisons. Since reconstruction requires support from two camera view minimum, rejecting a valid portion from one viewpoint could result in the loss of useful geometric information in another viewpoint. The three cases where accuracy was higher was likely caused because DINOv2 was able to remove false positives from empty space. By having these controlled conditions be part of the experiment, the additional filtering step could have introduced unnecessary restrictions rather than helping to improve reconstruction accuracy.

Across the 13 verified slicing comparisons, producing only the repair patch resulted in a median $93.08\%$ reduction in material usage along with a $89.39\%$ reduction in printing time. The median reduction in solid volume was $95.60\%$, as summarized in Table~\ref{tab:printing}.

\begin{table}[h]
\centering
\caption{Median reductions for 13 verified
paired slicing comparisons.}
\label{tab:printing}
\small
\renewcommand{\arraystretch}{1.15}
\begin{tabular}{@{}lr@{}}
\toprule
Measurement & Reduction \\
\midrule
Solid volume & 95.60\% \\
Estimated filament use & 93.08\% \\
Estimated printing time & 89.39\% \\
\bottomrule
\end{tabular}
\end{table}

This experiment used computer generated images with no noise and camera viewpoints that were already known. Because of this, we did not test how the system would perform with real camera images, inacurate camera positions, or imperfect depth measurement. We also only tested one type of simulated damage on the 20 components, meaning results may vary with other types of damage. Our findings with the implementation of artificial intelligence only apply in the specific methods in which they were implemented. Additionally, we didn't print the repair patches physically or test them in the real world. Future research could expand on these limitations by introducing new types of damaged, methods that better identify empty spaces hidden from the camera, and testing real world damaged components.

\FloatBarrier
\section{Conclusion}

This study used a pipeline to investigate whether missing sections of an automotive component could be reconstructed using paired images of the intact and synthetically damaged CAD models. Across the 160 experimental trials, 109 of them satisfied both accuracy requirements. The results show that camera coverage plays a vital role within the accuracy of the reconstruction, as the less viewpoints were available, the less accurate it was. In addition, the chosen implementation of DINOv2 did not improve the overall performance of the reconstruction and increased prediction time. The final evaluations were also able to show that printing time and material usage were significantly lowered when printing the repair patch rather than the entire part. Future work could touch on the reconstruction of hidden spaces, physical printing, and damaged real world objects.

\bibliographystyle{ieeetr}
\bibliography{sampleBib}

\end{document}